\documentclass[letterpaper, 10 pt, conference]{format/ieeeconf}  

\IEEEoverridecommandlockouts                              

\usepackage{dsfont}
\usepackage{amsmath}

\usepackage{graphics} 
\usepackage{amsmath} 
\usepackage{amssymb}  
\usepackage{todonotes}

\usepackage{amsfonts}
\usepackage{mathtools}
\usepackage[linesnumbered,algoruled,boxed,vlined, noend]{algorithm2e}
\usepackage{amsthm}
\usepackage{comment}
\usepackage{cite}
\usepackage{wrapfig}
\let\labelindent\relax
\usepackage{enumitem}
\usepackage{overpic}
\usepackage{hyperref}
\usepackage{dsfont}

\DeclareMathAlphabet{\mathcal}{OMS}{cmsy}{m}{n} 

\newtheorem{proposition}{Proposition}

\theoremstyle{definition}

\theoremstyle{remark}

\DeclarePairedDelimiterX{\norm}[1]{\lVert}{\rVert}{#1}

\newcommand{\rom}[1]{\uppercase\expandafter{\romannumeral #1\relax}}

\newcommand{\cspace}{\text{$\mathcal{Q}$}}

\newcommand{\arrangementspace}{\text{$\mathcal{A}$}}
\newcommand{\workspace}{\text{$\mathcal{W}$}}
\newcommand{\positionspace}{\text{$\mathcal{P}$}}
\newcommand{\manipulator}{\text{$\mathcal{M}$}}
\newcommand{\objects}{\text{$\mathcal{O}$}}

\newcommand{\DFSDP}{DFS_{DP}\xspace}

\newcommand{\CIDFSDP}{CIDFS_{DP}\xspace}

{\end{list}} 
\usepackage{soul} 
\usepackage[a-2b,mathxmp]{pdfx}[2018/12/22]

\usepackage{booktabs, makecell} 
\usepackage{multirow}

\title{\LARGE \bf
Efficient and High-quality Prehensile Rearrangement\\ in Cluttered and Confined Spaces}

\author{Rui Wang, Yinglong Miao, Kostas E. Bekris
\thanks{The authors are with the Department of Computer Science, Rutgers University, NJ, USA. Email: {\tt\small {\{rw485, ym420\}}@rutgers.edu} and {\tt\small kb572@cs.rutgers.edu}. The work is supported in part by an NSF HDR TRIPODS award 1934924.
}
}

\begin{document}

\maketitle
\thispagestyle{empty}
\pagestyle{empty}

\begin{abstract}
Prehensile object rearrangement in cluttered and confined spaces has broad applications but is also challenging. For instance, rearranging products in a grocery shelf means that the robot cannot directly access all objects and has limited free space. This is harder than tabletop rearrangement where objects are easily accessible with top-down grasps, which simplifies robot-object interactions. This work focuses on problems where such interactions are critical for completing tasks. It proposes a new efficient and complete solver under general constraints for monotone instances, which can be solved by moving each object at most once. The monotone solver reasons about robot-object constraints and uses them to effectively prune the search space. The new monotone solver is integrated with a global planner to solve non-monotone instances with high-quality solutions fast. Furthermore, this work contributes an effective pre-processing tool to significantly speed up online motion planning queries for rearrangement in confined spaces. Experiments further demonstrate that the proposed monotone solver, equipped with the pre-processing tool, results in 57.3\% faster computation and 3 times higher success rate than state-of-the-art methods. Similarly, the resulting global planner is computationally more efficient and has a higher success rate, while producing high-quality solutions for non-monotone instances (i.e., only 1.3 additional actions are needed on average). Videos of demonstrating solutions on a real robotic system and codes can be found at \href{https://github.com/Rui1223/uniform_object_rearrangement}{\color{blue}{https://github.com/Rui1223/uniform\_object\_rearrangement}}.
\end{abstract}

\section{Introduction}
\label{sec:intro}

Rearranging objects in confined spaces are often useful in logistic and domestic domains such as rearranging products in warehouse shelves and retrieving food from a packed refrigerator. At the same time, however, rearrangement in such confined spaces is more challenging than less confined setups, such as rearranging objects on a tabletop, which exhibit fewer constraints and allow for increased efficiency and providing desirable properties. In particular, tabletop settings allow the robot to reach the majority of objects at any point in time by using top-down grasps and then lift them above the other objects. This allows ignoring robot-object as well as object-object collisions during the rearrangement process. As a result, the only hard constraints arise from the potential overlap between the start and goal poses of objects.

\begin{figure}[t]
    \centering
    \includegraphics[width=0.49\textwidth]{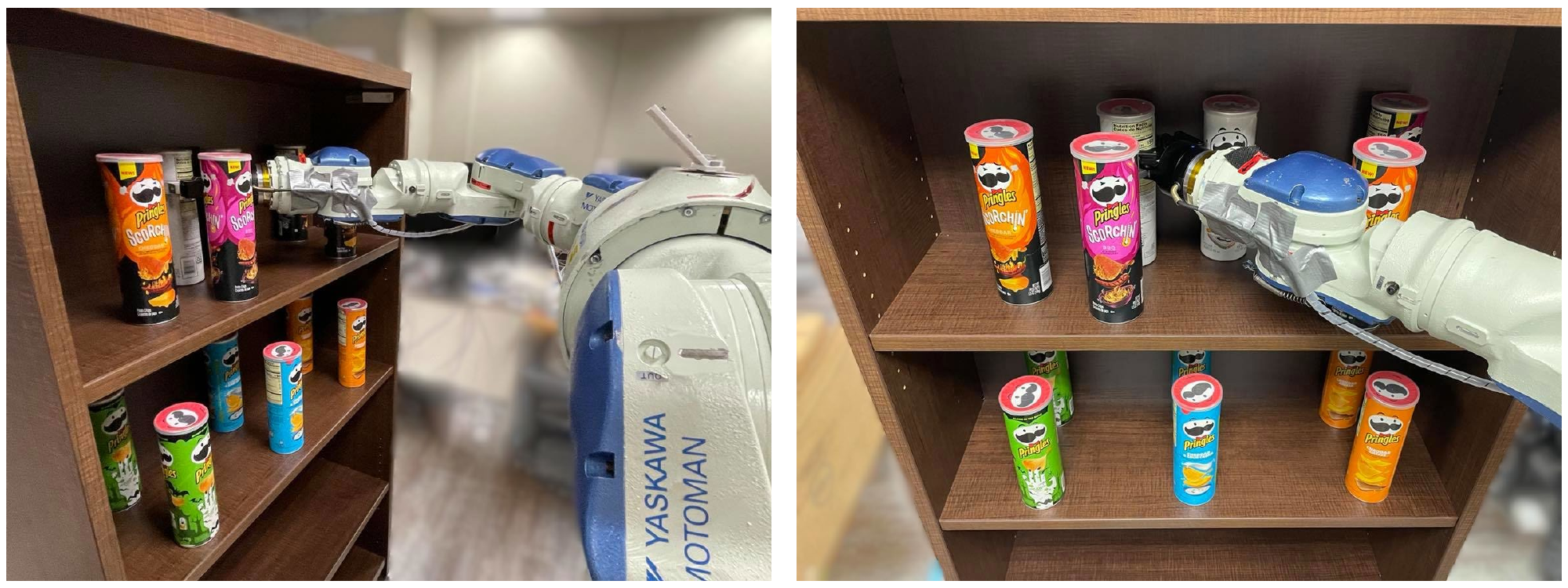}
    \vspace{-.25in}
    \caption{(left) An example rearrangement problem in a confined space. (right) The problem is challenging even with a few objects as the swept volume of the arm's motion can easily lead to collisions.}
    \label{fig:rearrangement_application}
\end{figure}

Recent work \cite{wang2021uniform} studied the tabletop rearrangement of uniformly-shaped cylinders, where object-object collisions cannot be ignored. It provided an efficient and complete monotone solver $\tt \DFSDP$ based on dynamic programming that outperformed backtracking (e.g., $\tt mRS$ \cite{stilman2007manipulation}). In \emph{monotone} instances, all objects can be moved at most once, while in \emph{non-monotone} instances, some objects have to be moved first to an intermediate position, i.e., a \emph{buffer}. The insight of $\tt \DFSDP$ is that instead of searching the space of all object permutations ($O(n!)$) to solve monotone instances, it is sufficient to search the space of object arrangements ($O(2^n)$). 

The first insight of the current paper is that even in the reduced arrangement space, there is a lot of redundancy and some branches of the search tree will not lead to a solution as they violate constraints enforced in cluttered and confined spaces. If these branches can be detected beforehand, they can be pruned to increase efficiency. Consequently, it is possible to improve upon the efficiency of $\tt \DFSDP$ while maintaining its completeness (which relies on the completeness of the underlying motion planner).

In addition, the current work focuses on the harder setups (Fig. \ref{fig:rearrangement_application}) where the arm must carefully maneuver to avoid both robot-object and object-object collisions. These constraints significantly increase computational cost due to more expensive collision checking. They also imply that the underlying motion planner can be at best probabilistically complete. This work aims to improve upon the efficiency of the motion planner, which significantly impacts overall performance since it is called multiple times by the rearrangement solvers.

This work also addresses non-monotone instances by integrating the  monotone solver with a global planner, which is a probabilistically complete non-monotone solver \cite{wang2021uniform} that explores the placement of objects in buffers. The setup here does not allow for buffers outside of the confined, cluttered workspaces, which is the hard case. In summary, this work focuses on rearranging uniformly-shaped cylindrical objects in confined, cluttered spaces and contributes:

\noindent 1. \textbf{A more efficient constraint informed monotone solver}, which detects branches of the underlying search tree that can be pruned without loss of completeness, i.e., the method is complete, if the underlying motion planner is complete. 
The proposed monotone solver is 57.3\% faster and provides 3 times higher success rate than two leading alternatives with similar completeness guarantees: $\tt mRS$ \cite{stilman2007manipulation} and $\tt \DFSDP$ \cite{wang2021uniform}. 

\noindent 2. \textbf{A high-quality non-monotone global planner}, which uses the monotone solver as a local planner to effectively solve non-monotone instances. The proposed global planner with the proposed monotone solver integrated has a much higher success rate and efficiency than counterparts while maintaining an equal level of high quality, i.e., the solutions need only 1.3 buffers on average.

\noindent 3. \textbf{An effective pre-processing tool}, which improves the efficiency of motion planning in cluttered rearrangement. Given a workspace discretization, the approach stores offline on a roadmap the sets of possible object target locations that result in a collision with the arm to avoid expensive online collision checking. The tool speeds up rearrangement solvers (proposed and compared) 49.1\% on average.

\section{Related Work}
\label{sec:related}

Object rearrangement relates to Navigation or Manipulation Among Movable Obstacles (NAMO \cite{chen1990practical, stilman2005navigation, van2009path, nieuwenhuisen2008effective} and MAMO \cite{ota2002rearrangement, stilman2007manipulation} respectively), which are computationally hard \cite{hopcroft1984complexity, wilfong1991motion, bereg2006lifting, stilman2008planning, halperin2020space} and more challenging in dynamic and uncertain environment \cite{kakiuchi2010working, wu2010navigation, levihn2014locally, han2019toward}. It also relates to Task and Motion Planning (TAMP) where a hierarchy is proposed to combine a high-level task planner and a low-level motion planner \cite{kaelbling2010hierarchical, garrett2020pddlstream}. This work generally follows such hierarchy but identifies task-specific constraints to effectively prune the action space so as to achieve improved efficiency and scalability.

Even for tabletop rearrangement, the problem is hard to solve optimally \cite{han2018complexity}. 
It is modeled as optimal matching over a directed graph for two arms \cite{shome2021fast}, or Answer Set Programming (ASP) for cluttered surfaces \cite{havur2014geometric, dabbour2019placement}. Minimizing buffer usage solves non-monotone problems more efficiently \cite{gao2021minimizing, gao2021fast}. Monte Carlo Tree Search (MCTS) \cite{song2020multi, labbe2020monte, huang2021visual} and deep learning \cite{yuan2018rearrangement, zeng2020transporter} have been applied on rearrangement. The above methods may not be always transferable to the harder, confined setup considered here. 

A fundamental strategy to solve monotone problems in the general case, including confined setups, is backtracking search \cite{stilman2007manipulation}, which is complete but does not scale well. An alternative involves constructing a dependency graph \cite{van2009centralized}, which describes the constraints between objects. If the graph has no cycles, the problem is monotone and topological sorting provides the solution. In principle, one has to consider all possible arm paths for picking and transferring an object to generate the true dependency graph, which can be computationally intractable. Minimum Constraint Removal (MCR) paths \cite{hauser2014minimum} have been used to construct an approximation of the dependency graph with few constraints in practice \cite{krontiris2016efficiently}. Once a monotone solver is available, it can be integrated into a global planner for solving more general, non-monotone problems \cite{krontiris2016efficiently, krontiris2015dealing}. 
This work provides an alternative way to reason about constraints and aims for high quality solutions and improved efficiency, without losing completeness.

A closely related problem is object retrieval. A fast and complete algorithm has been proposed to determine obstacles to be relocated before retrieving a target \cite{lee2019efficient}, with an extension that studies where to relocate obstacles \cite{cheong2020relocate}. Often the objective is to minimize the number of actions until the target is accessible \cite{danielczuk2019mechanical, nam2021fast}. Additional challenges include low visibility and observability due to object occlusions \cite{dogar2014object, xiao2019online, wang2020safe, bejjani2020occlusion}. In the problem considered here, every object has a target location, which is a more constrained objective.

Prehensile actions require good knowledge of objects' 6D pose \cite{wen2020robust}. Non-prehensile actions have been explored as they can simultaneously move multiple objects \cite{ben1998practical, huang2019large, pan2020decision, vieira2022persistent}, quickly declutter a scene and help minimize uncertainty \cite{dogar2012planning} under a conformant probabilistic planning formulation \cite{koval2015robust, anders2018reliably}. Though predicting the effect of pushing actions has been studied \cite{lynch1993estimating, huang2021dipn}, they are not as predictable as prehensile ones. In tasks where objects need to be safely placed (e.g., not dropping objects), prehensile actions are preferred.

\section{Problem Formulation}
\label{sec:problem_formulation}
Consider a cubic workspace $\workspace \subset \mathbb{R}^3$ with $n$ movable uniformly-sized cylinders $\objects=\{o_1, \cdots, o_n\}$, each of which can acquire a position $p \in \mathbb{R}^2$ by resting stably at the bottom surface of the workspace. An \emph{arrangement} $\alpha \in \arrangementspace$ is an assignment of $\objects$ to a set of object positions $\{p_1, \cdots, p_n\}$, where $\arrangementspace$ is the arrangement space. $\alpha[o_i]=p_i$ indicates that object $o_i$ is at position $p_i$ given the arrangement $\alpha$.


A robot arm $\manipulator$ is tasked to transfer one object at a time and can access $\workspace$ from only one side of the cubic space. The arm acquires a \emph{configuration} $q \in \cspace$ where $\cspace$ is the C-space of the arm. The swept volume of the robot at $q$ is denoted as $V(q)$. If the arm is grasping an object, the swept volume includes the object's volume given the grasp. $q(\alpha[o_i])$ defines a configuration where the arm can grasp $o_i$ at position $\alpha[o_i]$.  A \emph{manipulation path} $\pi_i: [0,1] \to \cspace$ for an object $o_i$ corresponds to a sequence of configurations that move object $o_i$ from one position to another. Such a path is \emph{valid} if no collision arises between $\bigcup_{t=0}^{1}V(\pi_i(t))$ with the boundary of $\workspace$ (excluding the open side) and the static obstacles. A manipulation path $\pi_i$ can be decomposed into a \emph{transit path} where the arm is approaching $o_i$ to be picked at $\alpha[o_i]$ and a \emph{transfer path} where the arm is transferring $o_i$ to a new position $p^{\prime}_i$. This will result in a new arrangement $\alpha^{\prime}$ where $\alpha^{\prime}[o_i] = p_i^{\prime}$ and $\forall j \in \{1,\cdots,n\}, j \neq i: \alpha^{\prime}[o_j] = \alpha[o_j]$.

Based on these notations, the problem is defined as: given an initial arrangement $\alpha_I$ and a final arrangement $\alpha_F$ of $n$ objects $\objects$, find a sequence of valid manipulation paths $\Pi = (\pi_0, \pi_1, \ldots, )$, which moves all objects from $\alpha_I$ to $\alpha_F$. A problem is \emph{monotone} if the sequence $\Pi$ consists of at most one manipulation path for each object. Otherwise, the problem is \emph{non-monotone} and at least one object needs to be moved to a \emph{buffer} before being moved to its goal.

\emph{Assumptions}: The objects are cylinders of uniform, known size. Since cylinders are symmetric and the height ($z$ value) is known, the planar position $(x,y) \in \mathbb{R}^2$ of an object's center is sufficient to generate grasps for picking and placing the object by using an inverse kinematic (IK) solver. For motion planning and collision checking during the manipulation trajectory, however, the 6D poses of objects are used in order to define their swept volume in the 3D workspace.

\section{Methodology}
\label{sec:methodology}

The solution sequence $\Pi$ can be obtained by concatenating multiple manipulation paths, the order of which is selected by a rearrangement task planner. Each manipulation path is the result of calling a motion planner. The task planner itself is hierarchical, where a monotone solver attempts first to connect monotonically $\alpha_I$ and $\alpha_F$. 
If it fails, it returns a set of reachable arrangements from $\alpha_I$ as a partial solution to be used by the global planner. Then the global planner explores the selection of objects to be moved to buffers. Therefore, the solution quality is determined by all these three components: the global planner, the local monotone solver and the lower-level motion planner. This section covers all three aspects, starting with the monotone solver. 

\begin{figure}[ht]
    \centering
    \includegraphics[width=0.42\textwidth]{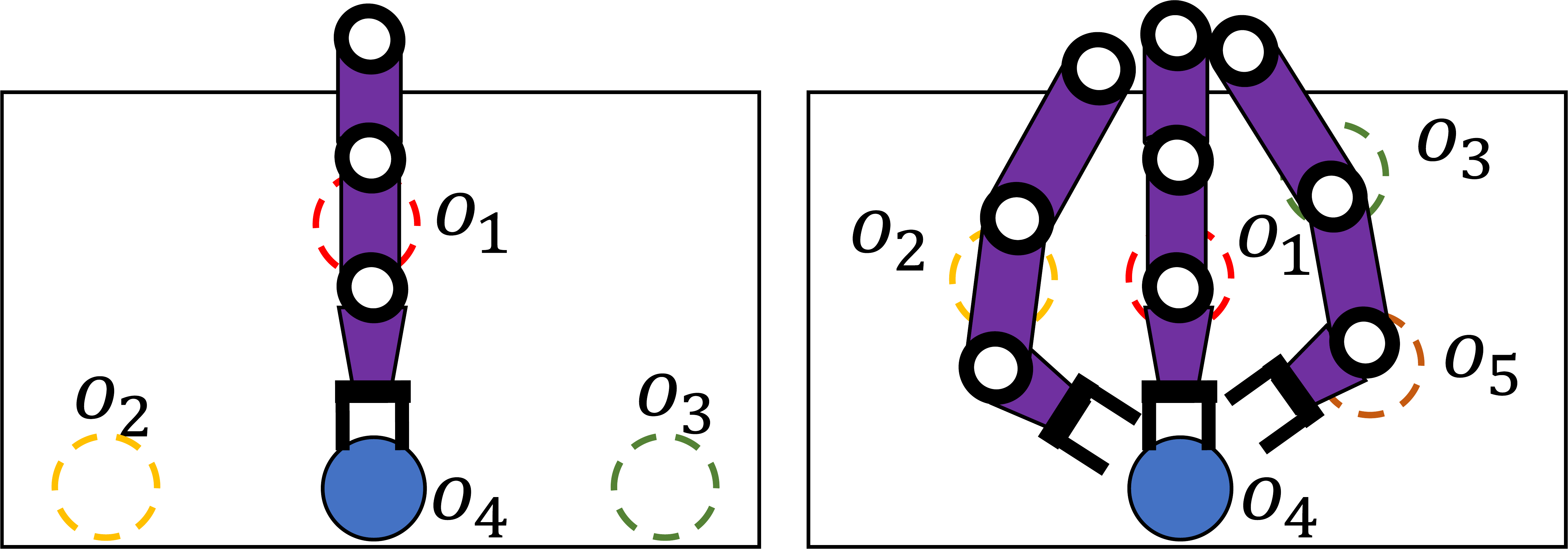}
    \vspace{-.05in}
    \caption{Two examples of the robot arm moving object $o_4$ (blue shaded circle). For simplicity of illustration, only the goal positions of other objects are shown (dashed circles). (left) The goal position of $o_1$ makes it impossible to arrange $o_4$ after the placement of $o_1$. (right) Similarly, the three arm configurations for grasping $o_4$ intersect with goal positions of $\{o_2\}, \{o_1\}$ and $\{o_3, o_5\}$, respectively.}
        \vspace{-.05in}
    \label{fig:illustrative_example}
\end{figure}

\subsection{Efficient and Complete Local Monotone Solver:\\ Constraint Informed Rearrangement Search (CIRS)}

Two methods which solve monotone problems are backtracking search, referred to here as $\tt mRS$ \cite{stilman2007manipulation}, with time complexity $O(n!)$ and dynamic programming, referred to here as $\tt \DFSDP$ \cite{wang2021uniform}, with time complexity $O(2^n)$. They are complete if the underlying motion planner is also complete.  



\begin{figure}[ht]
    \centering
    \includegraphics[width=0.47\textwidth]{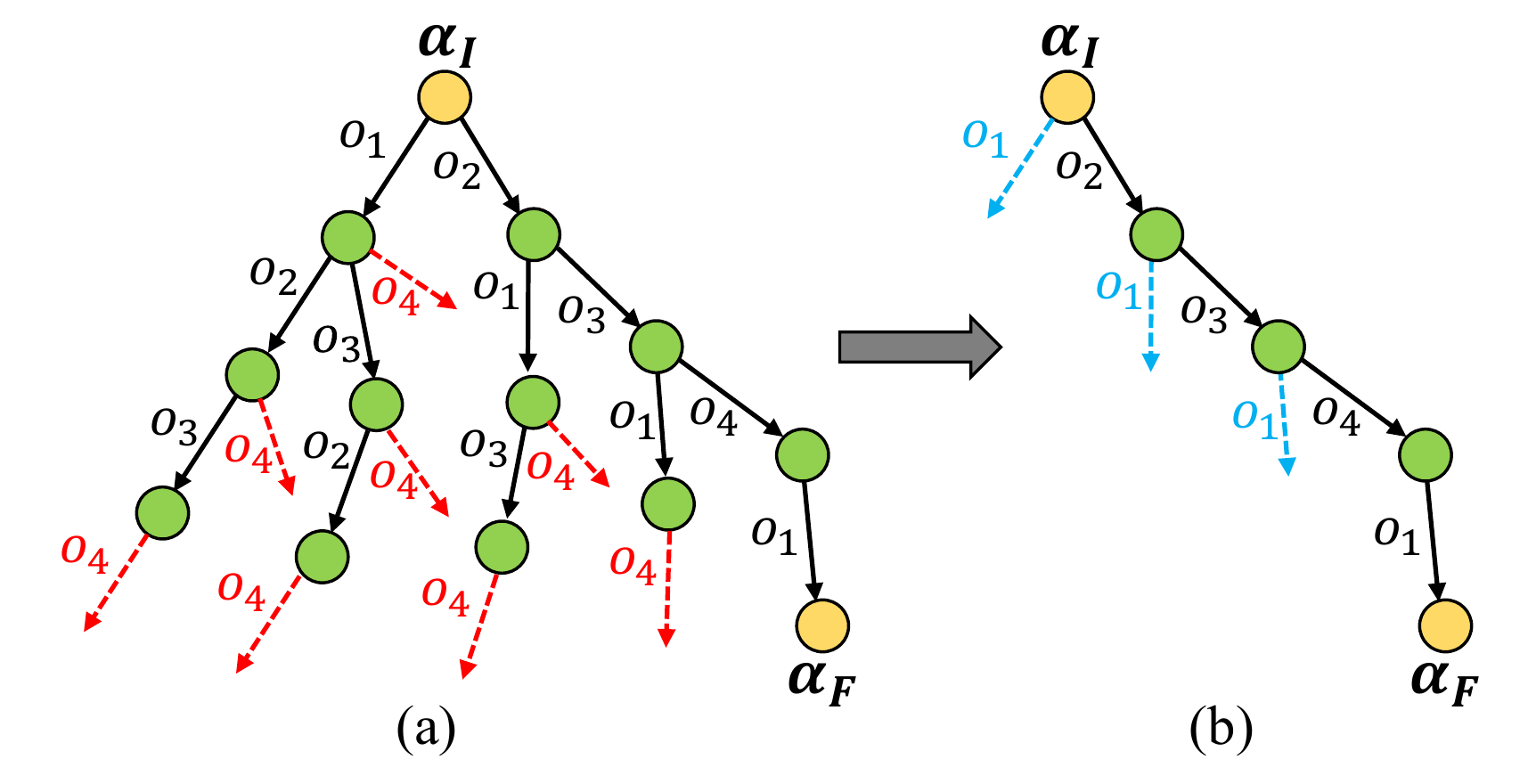}
        \vspace{-.1in}
    \caption{(a) A search tree that considers the sequence of actions to connect $\alpha_I$ to $\alpha_F$ (yellow circles) for the problem in Fig. \ref{fig:illustrative_example}(a). Each node (green or yellow circles) represents an arrangement $\alpha$ and each edge (arrows) indicates the transition between nodes. Black arrows indicate successful transitions and red ones unsuccessful. The solution is found as: $o_2 \rightarrow o_3 \rightarrow o_4 \rightarrow o_1$. (b) A search tree after enforcing the constraint of not moving $o_1$ while $o_4$ is at the start. Cyan arrows indicate the pruned actions.}
    \label{fig:search_process}
\end{figure}

Fig. \ref{fig:illustrative_example}(a) shows an example where the goal of $o_1$ hinders the rearrangement of $o_4$. A search process is demonstrated in Fig. \ref{fig:search_process}(a). The red arrows indicate the inability of rearranging $o_4$ after $o_1$, which is instantly detected upon arranging $o_4$. 
However, such inability should be detected in an earlier stage when arranging $o_1$ to its goal while $o_4$ is at its current position, as all branches afterwards will result in failure.
Therefore, the search should not consider moving $o_1$ at any arrangement states $\widetilde{\arrangementspace} = \{ \alpha \mid \alpha[o_4] = \alpha_C[o_4] \}$, where $\alpha_C$ represents the current arrangement state. Given this observation, if such invalid action can be detected at $\widetilde{\arrangementspace}$, the search tree can be significantly pruned, as in Fig. \ref{fig:search_process}(b).

A problem can be more involved as shown in Fig. \ref{fig:illustrative_example}(b) where each configuration for grasping $o_4$ intersects with the goals of a set of objects. Denote $\mathcal{C}_4^j$ as the colliding object set, where the goals of these objects hinder the grasping of $o_4$ for the $j$-th configuration. In this example, $\mathcal{C}_4^1 = \{o_2\}, \mathcal{C}_4^2 = \{o_1\}, \mathcal{C}_4^3 = \{o_3, o_5\}$. 
The cross product of the colliding object sets 
$\mathcal{C}_4^1\times\mathcal{C}_4^2\times\mathcal{C}_4^3$ results in two constraint sets $c^1 = \{o_1, o_2, o_3\}$ and $c^2 = \{o_1, o_2, o_5\}$, indicating that if objects in $\{o_1, o_2, o_3\}$ or $\{o_1, o_2, o_5\}$ are all moved to their goals before $o_4$, the move of $o_4$ will fail.

Therefore, each constraint set can elicit a set of arrangements $\widetilde{\arrangementspace}$ where moving certain object becomes invalid. Take $c^1 = \{o_1, o_2, o_3\}$ as an example, from which it is true that moving $o_1$ is invalid at any arrangement states \[\widetilde{\arrangementspace} = \{ \alpha \mid \alpha[o_4] = \alpha_C[o_4], \alpha[o_2] = \alpha_F[o_2], \alpha[o_3] = \alpha_F[o_3] \}\]
Similar observations can be made on moving $o_2$ and $o_3$.

It now becomes clear that it is beneficial to build a structure $\mathds{A}_{invalid}: \objects \rightarrow \widetilde{\arrangementspace}$ to store all invalid actions of moving an object $o \in \objects$ at an arrangement state $\alpha \in \widetilde{\arrangementspace}$. $\mathds{A}_{invalid}$ is then used to prune the search tree by disallowing invalid actions.
The two examples are generalized as the Constrained Informed Rearrangement Search ($\tt CIRS$) approach, which is the monotone solver and shown in Alg. \ref{alg:CIRS} with two steps (1) detection of invalid actions upon arrangements before the search (line 1) and (2) a search process with informed constraints obtained from step 1 (line 2).

\begin{algorithm}
\label{alg:CIRS}
\DontPrintSemicolon
\begin{small}
    \SetKwInOut{Input}{Input}
    \SetKwInOut{Output}{Output}
    \SetKwComment{Comment}{\% }{}
    \caption{$\tt CIRS$($\alpha_C$, $\alpha_F$, \objects, $K$)}
    \SetAlgoLined
        $\mathds{A}_{invalid}$ = $\textsc{DetectInvalidity}(\alpha_C, \alpha_F, \objects, K)$\\
        {\bf return} $\ \tt \CIDFSDP$($ \emptyset, \alpha_C, \alpha_F$, $ \mathds{A}_{invalid}$)\\
\end{small}
\end{algorithm}

Step 1 is described in Alg. \ref{alg:DetectInvalidity}, which constructs $\mathds{A}_{invalid}$. For each object $o_i$ (line 3), $K$ grasping configurations $[q^1, \cdots, q^K]$ are generated for moving $o_i$ from $\alpha_C$ to $\alpha_F$ (line 4). Here $q^k$ is short for $q^k(\alpha_C[o_i])$ or $q^k(\alpha_F[o_i])$ representing the $k^{th}$ arm configuration either to pick $o_i$ at $\alpha_C$, or to place $o_i$ at $\alpha_F$. As mentioned in the example (Fig. \ref{fig:illustrative_example}(b)), the cross product of all colliding object sets $\{\mathcal{C}_{i}^{1} \times \cdots \times \mathcal{C}_{i}^{K}\}$ gives all constraint sets (line 5), each of which (denoted as $c^j$) elicits a set of arrangements $\widetilde{\arrangementspace}$ (line 6) to be added to $\mathds{A}_{invalid}$ (line 7).
\begin{algorithm}
\label{alg:DetectInvalidity}
\DontPrintSemicolon
\begin{small}
    \SetKwInOut{Input}{Input}
    \SetKwInOut{Output}{Output}
    \SetKwComment{Comment}{\% }{}
    \caption{$\textsc{DetectInvalidity}$($\alpha_C$, $\alpha_F$, \objects, $K$)}
    \SetAlgoLined
        \For{$o_i \in \objects$}{
            $\mathds{A}_{invalid}[o_i] = \emptyset$\\}
        \For{$o_i \in \objects$}{
            $[q^1, \cdots, q^K], [\mathcal{C}_{i}^{1}, \cdots, \mathcal{C}_{i}^{K}]=$ $\textsc{GenerateArmConfigurations}$($\alpha_C, \alpha_F, K$)\\
            \For{$c^j \in \{\mathcal{C}_{i}^{1} \times \cdots \times \mathcal{C}_{i}^{K}\}$}{
                $\widetilde{\arrangementspace}=$ $\textsc{ElicitArrangements}$($c^j$)\\
                $\mathds{A}_{invalid}$.\textsc{add}($\widetilde{\arrangementspace}$)
            }
        }
        \Return $\mathds{A}_{invalid}$\\
\end{small}
\end{algorithm}


Once $\mathds{A}_{invalid}$ is constructed, step 2 is performed for searching the solution for the monotone problem $\alpha_C \rightarrow \alpha_F$, which is shown in Alg. \ref{alg:CIDFSDP}. $\tt \CIDFSDP$ is a search method recursively solving a subproblem: $\alpha_C \rightarrow \alpha_F$, which is built on top of the original $\tt \DFSDP$ \cite{wang2021uniform}. It grows a search tree $T$ in the arrangement space $\arrangementspace$ from $\alpha_C$ to $\alpha_F$. Every time an object yet to move $o$ is selected at $\alpha_C$ (line 1), 
Alg. \ref{alg:CIDFSDP} checks if $\alpha_C$ is one of the arrangements that invalidates the action of moving object $o$ (line 2). 
If it does, Alg. \ref{alg:CIDFSDP} will not consider moving $o$ at $\alpha_C$ and moves on to another object (line 2). 
Otherwise, moving $o$ is valid. If the resulting $\alpha_{new}$ (line 3-4) after moving $o$ has not been explored before (line 5), a motion planner is called to check the feasibility of moving $o$ from $\alpha_C[o]$ to $\alpha_F[o]$ (line 6). If a feasible $\pi$ is found (line 7), the search tree will expand to $\alpha_{new}$ (line 8). If the problem is not solved yet (line 9), a recursive process will be triggered to solve a subproblem $\alpha_{new} \rightarrow \alpha_{F}$ (line 10). If the problem is solved (line 11), the search tree $T$ is returned, from which the path sequence $\Pi$ can be obtained. Otherwise, a subtree is returned as a partial solution (line 12).

\begin{algorithm}
\label{alg:CIDFSDP}
\DontPrintSemicolon
\begin{small}
    \SetKwInOut{Input}{Input}
    \SetKwInOut{Output}{Output}
    \SetKwComment{Comment}{\% }{}
    \caption{$\tt \CIDFSDP$($T$, $\alpha_C$, $\alpha_F$, $\mathds{A}_{invalid}$)}
    \SetAlgoLined
        \For{$o\in \objects\backslash\objects(\alpha_C)$}{
            \lIf{$\alpha_{C} \in \mathds{A}_{invalid}[o]$}{continue}
            $\alpha_{new}[\objects\backslash \{o\}]=\alpha_{C}[\objects\backslash \{o\}]$\\
            $\alpha_{new}[o]=\alpha_{F}[o]$\\
            \If{$\alpha_{new} \notin T$}{
                $\pi \leftarrow$ $\textsc{MotionPlanning}$($\alpha_C$, $\alpha_{new}$, $o$)\\
                \If{$\pi \neq \emptyset$}{
                    $T[\alpha_{new}].parent\leftarrow \alpha_{C}$\\ 
                    \If{$\alpha_{new} \neq \alpha_F$}{
                        $T$ = $\tt \CIDFSDP$($T$, $\alpha_{new}$, $\alpha_F$, $\mathds{A}_{invalid}$)\\
                    }
                    \lIf{$\alpha_F \in T$}{\Return $T$}
                }
            }
        }
        \Return $T$\\
\end{small}
\end{algorithm}

\vspace{-0.05in}
The key feature of the monotone solver is that it checks whether it is valid to move an object $o$ at the current arrangement $\alpha_C$ (line 2). If it is not, Alg. \ref{alg:CIDFSDP} will not generate a new arrangement $\alpha_{new}$. Therefore, no time will be wasted on growing a useless subtree rooted at such $\alpha_{new}$, which gives significant speed-ups as shown in section \ref{sec:experiments}.


\vspace{-.05in}
\begin{proposition}
CIRS is complete for any monotone rearrangement problem  given a generalized constraint checker and a complete motion planner.
\end{proposition}
\vspace{-.2in}
\begin{proof}
It suffices to show that if the problem has at least one solution, CIRS will return one. W.l.o.g, denote one of the solutions as $\Pi=(\pi_1,\dots,\pi_n)$ corresponding to an object arrangement order $[o_1,\dots,o_n]$. CIRS only prunes arrangement sequences that violate the constraints as shown in the generation of $\mathds{A}_{invalid}$ and in the constraint checker. Meanwhile, CIRS exhaustively searches for all possible objects at each step that have not been arranged yet. Given the completeness of the motion planner, the search tree includes the solution sequence $[o_1,\dots,o_n]$ and feasible motion plans can be found to execute it. \vspace{-.05in}
\end{proof}

\subsection{Addressing Non-Monotone Challenges: (PERTS)}
When a problem cannot be solved monotonically, the local monotone solver can return a partial solution, i.e., a subtree of arrangements attached to $\alpha_I$. The proposed approach follows a systematic way of building a global tree out of these partial trees. It selects an existing arrangement that is reachable from the root node $\alpha_I$ and performs an action where an object moves to a buffer. Such an action is referred to here as a \emph{perturbation} and is a feature of the global planner. Alg. \ref{alg:PERTS} describes how perturbations are used in the global task planner $\tt PERTS$ (short for perturbation) to extend beyond monotonically reachable arrangements, which are generated by $\tt CIRS$.


\begin{algorithm}
\label{alg:PERTS}
\DontPrintSemicolon
\begin{small}
    \SetKwInOut{Input}{Input}
    \SetKwInOut{Output}{Output}
    \SetKwComment{Comment}{\% }{}
    \caption{$\tt PERTS(CIRS)$($\alpha_I$, $\alpha_F$, \objects)}
    \SetAlgoLined
        $T = \emptyset$, $\Pi = \emptyset$\\
        $T_{sub} = {\tt CIRS}(\alpha_I, \alpha_F, \objects, K)$\\
        $T = T + T_{sub}$\\
        \While{$\alpha_F \notin T$ and \textsc{TimePermitted}}{
            $\alpha_C = \textsc{SelectNode}(T)$\\
            $\alpha_{pert} = \textsc{PerturbNode}(\alpha_C)$\\
            \lIf{$\alpha_{pert} = \emptyset$}{continue}
            $T_{sub} = {\tt CIRS}(\alpha_{pert}, \alpha_F, \objects, K)$\\
            $T = T + T_{sub}$\\
        }
        \If{$\alpha_F \in T$}{
            $\Pi = \textsc{TraceBackPath}(T, \alpha_F, \alpha_I)$\\
        }
        \Return $\Pi$\\
\end{small}
\end{algorithm}

\begin{figure}[ht]
    \centering
    \includegraphics[width=0.40\textwidth]{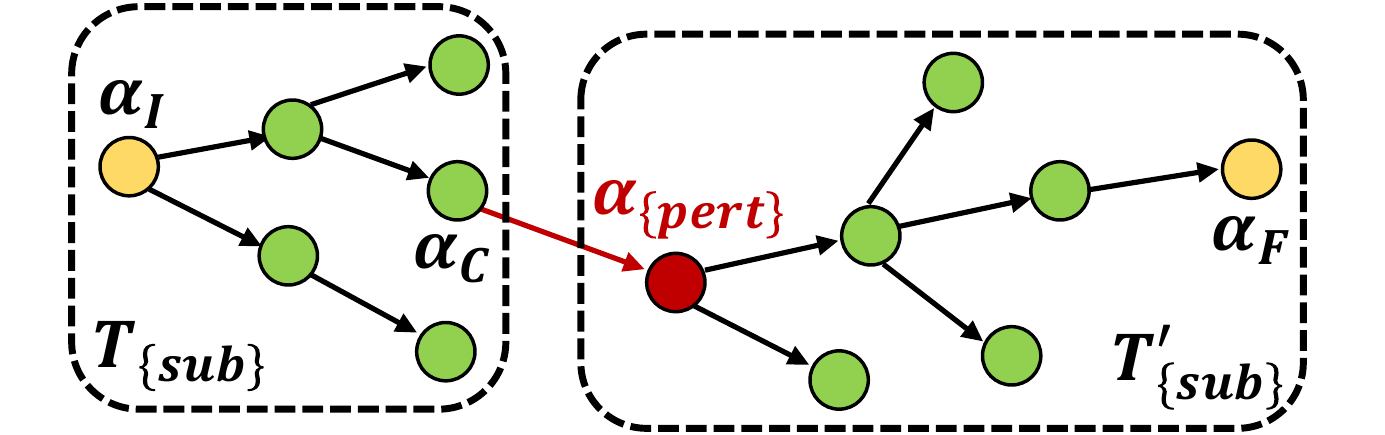}
    \vspace{-.05in}
    \caption{The global search tree from $\tt PERTS$. The monotone solver is used to connect $\alpha_I$ to $\alpha_F$ (yellow circles). If this fails, it can still provide a subtree $T_{sub}$ out of $\alpha_I$. Then, a node $\alpha_C$ on $T_{sub}$ is selected to perform a perturbation (red arrow), i.e., an object is moved to a buffer. This leads to a node $\alpha_{pert}$ (red circle). Then, the monotone solver is called to connect $\alpha_{pert}$ to $\alpha_F$. If the resulting subtree $T_{sub}^{'}$ connects with $\alpha_{F}$, the non-monotone problem is solved with a single buffer.}
    \label{fig:perts_global_tree}
\end{figure}

Given the task of rearranging objects $\objects$ from $\alpha_I$ to $\alpha_F$, the local solver first tries to solve the problem with a monotonic connection (lines 1-3). If the problem is solved in a given time, the solution path sequence $\Pi$ can be obtained by tracing back the path along the search tree $T$ (line 10-11). Otherwise (line 4), a node $\alpha_C$ from the subtree reachable from the root $\alpha_I$ is randomly selected to perform a random perturbation (randomly select an object to be placed in a buffer randomly selected) (line 5-6). If such perturbation is not successful (line 7), line 5 and 6 will be repeated given time permitted. Otherwise, such perturbation results in perturbation node $\alpha_{pert}$. The monotone solver then is called to solve the task of monotonically rearranging from $\alpha_{pert}$ to $\alpha_F$ (line 8-9). This process continues until either a solution is eventually found (line 10-11) or time exceeds a specified threshold (line 4). Fig. \ref{fig:perts_global_tree} illustrates the process. This search can also be executed in a bidirectional manner but here for simplicity, the unidirectional version is described and used.

\begin{figure*}[ht]
    \centering
    \includegraphics[width=0.80\textwidth]{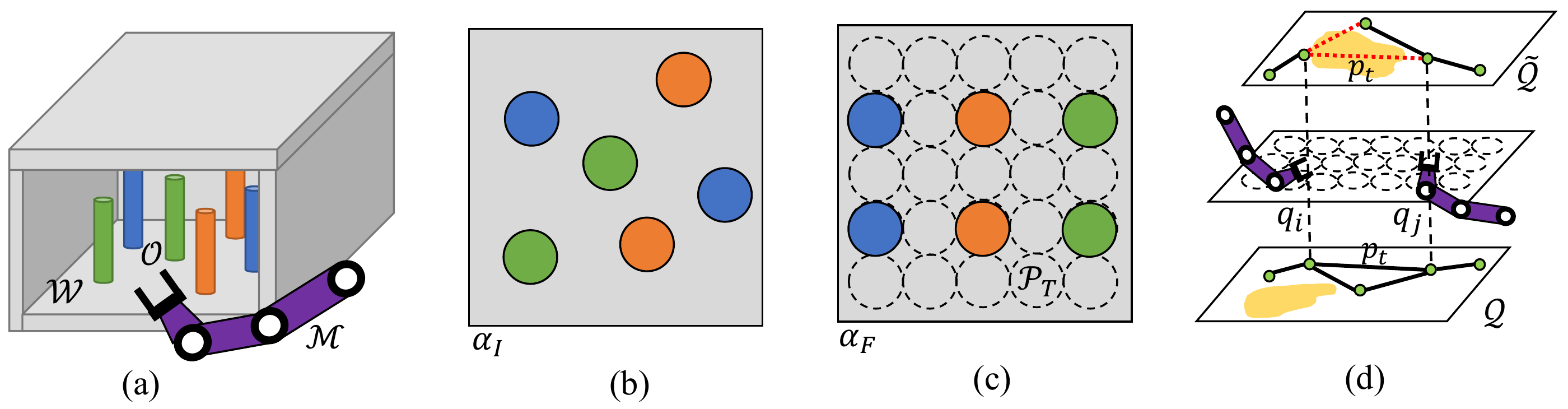}
    \vspace{-.1in}
    \caption{(a) An example of a robot arm $\mathcal{M}$ rearranging objects $\objects$ in the workspace $\mathcal{W}$. (b) The current positions of all 6 objects in the workspace. (c) The workspace is discretized offline given a dense set of candidate positions for the objects. In this task, goal positions are selected so as to rearranging objects to be aligned in two rows. (d) Offline collision checking is performed on each transition from one arm configuration to another based on all pre-defined object positions (middle layer). For instance, if the transition from $q_i$ to $q_j$ collides with an object position $p_t$, the edge connecting $q_i$ and $q_j$ will be labeled with $p_t$ (top and bottom layer). Here the roadmap consists of two modes where the validity of each edge can vary (1) $\mathcal{Q}$: transiting to an object (bottom layer) (2) $\widetilde{\mathcal{Q}}$: transferring an object (top layer). For instance, transition from $q_i$ to $q_j$ is valid (black solid line) in $\mathcal{Q}$ (bottom layer) while it is not invalid (red dotted line) in $\widetilde{\mathcal{Q}}$ (top layer) as the grasped object results in a collision (yellow regions)}
    \label{fig:labeled_roadmap}
\end{figure*}

The \emph{perturbation level} of an arrangement $\alpha$ is defined as the total number of perturbations (i.e., buffers) it takes to reach $\alpha$ from $\alpha_I$. Alg. \ref{alg:PERTS} increments the perturbation level one at a time and grows a subtree $T_{sub}$ rooted at $\alpha_{pert}$, which consists of nodes with the same perturbation level as that of $\alpha_{pert}$.  In this way, $\tt PERTS$ promotes high-quality solutions in terms of low number of buffers used to fulfill the task.

\subsection{Speeding Up Motion Planning: Labeled Roadmap}

Efficiently computing whether transitions between arrangement nodes are feasible plays a critical role in solving the task efficiently. This relates to the computational efficiency of the underlying motion planner. A sampling-based planner similar to $\tt PRM^*$ is used to first generate a roadmap upon which any standard search algorithms, such as $\tt A^*$, can be used to search. Due to sampling-based nature of the approach, the quality of the solution sometimes may suffer. Furthermore, the cost of online query resolution can be high due to the amount of collision checking required in confined and cluttered spaces.

The problem requires the swept volume of the arm's motion to be minimized inside the confined workspace to minimize the chances of colliding with the objects. This work generates a roadmap with a significant ratio of the arm configurations ($50\%$ in experiments) capable of grasping objects at different reachable positions inside the workspace. The remaining set of configurations are randomly sampled to provide coverage of the C-space. With this pre-processing, the pick and place configurations are more likely to connect to each other. They can also produce shorter solutions paths, which minimize the chances of intersecting objects.

This work also incorporates offline collision information on each roadmap edge, which corresponds to the transition between two sampled arm configurations. The purpose is to save computation from collision checking online. To achieve that, the workspace is discretized into a set of possible object positions $\positionspace_{T}$. Then the objects' goal positions can be selected from the pre-processed set $\positionspace_{T}$ (Fig. \ref{fig:labeled_roadmap}(c)) but the start positions do not have to be aligned with the pre-processing (Fig. \ref{fig:labeled_roadmap}(b)). For instance, consider a practical scenario where a grocery store customer casually leaves an item they no longer want to purchase in a shelf and the task is to put the item back to a pre-assigned grid location. 

Given the pre-processing, the most expensive part of the collision checking can be performed offline on each edge in the roadmap (Fig. \ref{fig:labeled_roadmap}(d)) to detect if robot-object or object-object collisions arise assuming an object is at position $p_t \in \positionspace_{T}$. If a collision occurs, the edge will be labeled with that corresponding position $p_t$. During online planning, if the planning query takes place at an arrangement $\alpha$ where position $p_t$ is occupied, that edge will not be considered by the A* search algorithm when it is called at $\alpha$. This process also differentiates two manipulation modes, one for the arm transiting to an object, and one for transferring to an object. This pre-processing results in significant speed-ups, as will be shown in section \ref{sec:experiments}. 

The labeled roadmap does not incorporate collision information for the initial positions of the objects, which are assigned online upon the generation of an instance. In order to utilize the labeled roadmap, each initial object position is approximated by the nearest pre-defined position, which is then used for online planning query. This results in an approximation that depends on the density of the discretization and affects completeness. In that regard, the pre-processing introduces a level of resolution completeness.

\section{Experiments}
\label{sec:experiments}

This section evaluates the effectiveness and the impact of the proposed work: (1) pre-processed labeled roadmap; (2) efficient local monotone solver $\tt CIRS$; and (3) global task planner $\tt PERTS$.


\begin{table*}[t!]
\parbox[t][][b]{.67\linewidth}{
\begingroup
\setlength{\tabcolsep}{0pt} 
\centering
\begin{tabular}{@{\extracolsep{2pt}}ccccccc@{}}
\toprule
\multirow{2}{*}{\# Objects}&\multicolumn{2}{c}{CIRS}&\multicolumn{2}{c}{$\textrm{DFS}_\textrm{DP}$}&\multicolumn{2}{c}{mRS}\\
\cmidrule{2-3}\cmidrule{4-5}\cmidrule{6-7}
&\multicolumn{1}{c}{w/o}&\multicolumn{1}{c}{w/}&\multicolumn{1}{c}{w/o}&\multicolumn{1}{c}{w/}&\multicolumn{1}{c}{w/o}&\multicolumn{1}{c}{w/}\\
\midrule
    $7$ & 
    $23.2\textrm{ } (100\%)$ & 
    $\boldsymbol{8.9}\textrm{ } (\boldsymbol{100\%})$ & 
    $40.5\textrm{ }(100\%)$ &
    $\boldsymbol{19.5}\textrm{ }(\boldsymbol{100\%})$ &
    $68.2\textrm{ }(83\%)$ &
    $\boldsymbol{45.8}\textrm{ }(\boldsymbol{92\%})$\\

    $8$ & 
    $27.1\textrm{ } (100\%)$ & 
    $\boldsymbol{11.2}\textrm{ } (\boldsymbol{100\%})$ & 
    $83.8\textrm{ }(80\%)$ &
    $\boldsymbol{37.5}\textrm{ }(\boldsymbol{100\%})$ &
    $112.2\textrm{ }(60\%)$ &
    $\boldsymbol{80.8}\textrm{ }(\boldsymbol{73\%})$\\

    $9$ & 
    $75.9\textrm{ }\textrm{ } (88\%)$ & 
    $\boldsymbol{18.2}\textrm{ } (\boldsymbol{100\%})$ & 
    $121.8\textrm{ }(65\%)$ &
    $\boldsymbol{53.9}\textrm{ }(\boldsymbol{91\%})$ &
    $141.5\textrm{ }(48\%)$ &
    $\boldsymbol{99.0}\textrm{ }(\boldsymbol{65\%})$\\

    $10$ & 
    $58.4\textrm{ } (100\%)$ & 
    $\boldsymbol{19.1}\textrm{ } (\boldsymbol{100\%})$ & 
    $156.1\textrm{ }(57\%)$ &
    $\boldsymbol{80.9}\textrm{ }(\boldsymbol{91\%})$ &
    $173.0\textrm{ }(21\%)$ &
    $\boldsymbol{132.9}\textrm{ }(\boldsymbol{43\%})$\\

\bottomrule
\end{tabular}
\caption{Comparison with and without the labeled roadmap: Seconds/ (Success Rate)}
\label{table:labeled_roadmap_results}
\endgroup}
\parbox[t][][b]{.32\linewidth}
{
\centering
\begingroup
\setlength{\tabcolsep}{2pt} 
\begin{tabular}{cccc}
\toprule
\# Objects & CIRS & $\textrm{DFS}_{\textrm{DP}}$ & mRS \\
\midrule
    $9$ & 
    $\boldsymbol{1.0}\textrm{ } (\boldsymbol{100\%})$ & 
    $1.3\textrm{ } (75\%)$    & 
    / $(0\%)$ \\
    $10$ &
    $\boldsymbol{1.2}\textrm{ } (\boldsymbol{85.7\%})$ &  
    $1.0\textrm{ } (14.3\%)$ & 
    / $(0\%)$ \\

    $11$ &
    $\boldsymbol{1.6}\textrm{ } (\boldsymbol{94.1\%})$ &  
    $1.5\textrm{ } (11.8\%)$ & 
    / $(0\%)$ \\

    $12$ &
    $\boldsymbol{1.5}\textrm{ } (\boldsymbol{88.2\%})$ &  
    / $(0\%)$ & 
    / $(0\%)$ \\

\bottomrule
\end{tabular}
\caption{Number of buffers needed for non-monotone instances (Success Rate)}
\label{table:actions_non_monotone}
\endgroup}
\end{table*}


{\bf Impact of Pre-processing:} The proposed pre-processing is first evaluated in terms of the speed-up it provides to the low-level motion planner. Here the experiments are performed with and without the labeled roadmap (random sampling + unlabeled edges) on the monotone problems using the proposed $\tt CIRS$ and the comparison methods $\tt mRS$ and $\tt \DFSDP$ given a limit of 3 minutes to find a solution. The number of samples in the roadmap is 2000. 7-10 objects are selected to test the effectiveness of the proposed labeled roadmap in speeding up and increasing feasibility under the given time threshold. 30 experiments are performed on each number of objects. Table \ref{table:labeled_roadmap_results} demonstrates the performance difference of all methods (proposed and comparison methods) when they are implemented with (w/) and without (w/o) the labeled roadmap. The proposed labeled roadmap provides 65.9\%, 52.8\% and 28.5\% speed-ups for $\tt CIRS$, $\tt \DFSDP$ and $\tt mRS$, respectively. Furthermore, introducing the labeled roadmap improves the  feasibility of solving harder problems for less efficient methods, as the success rate increases from 57\% to 91\% for $\tt \DFSDP$ and from 21\% to 43\% for $\tt mRS$ on 10-object cases. Based on this computational improvement and to elicit the best performance for all the methods in the following experiments on monotone and non-monotone problems, all methods are compared with the labeled roadmap integrated.

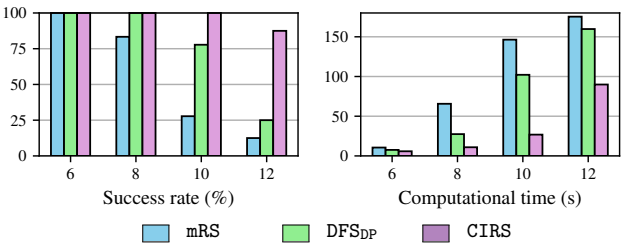
\begin{figure}[ht]
    \begin{center}
        \begin{overpic}[scale=0.68]{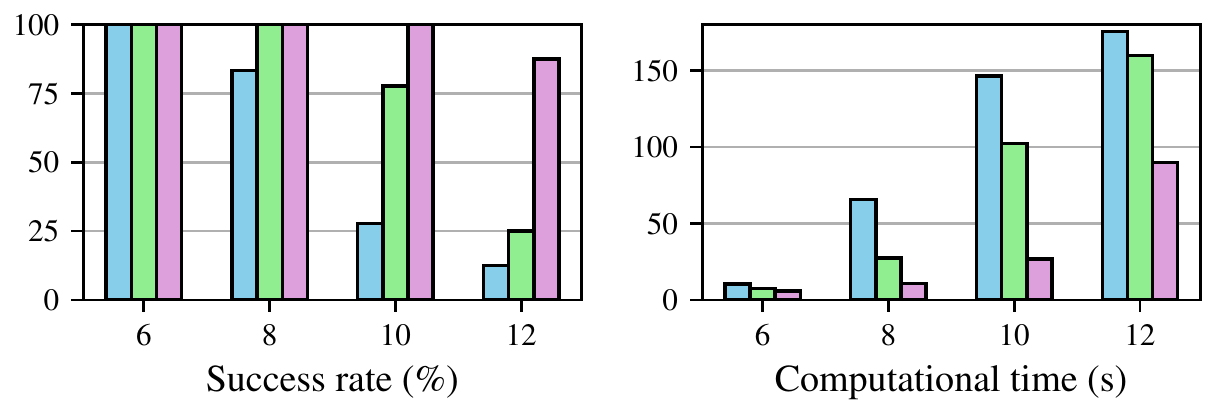}
        \definecolor{skyblue}{RGB}{135, 206, 235}
        \put(23,-3){\fcolorbox{black}{skyblue}{  }}
        \put(30, -3.5){{\footnotesize $\tt mRS$}}
        \definecolor{lightgreen}{RGB}{144, 238, 144}
        \put(45,-3){\fcolorbox{black}{lightgreen}{  }}
        \put(52, -3.5){{\footnotesize $\tt \DFSDP$}}
        \definecolor{plum}{RGB}{178, 132, 190}
        \put(67,-3){\fcolorbox{black}{plum}{  }}
        \put(74, -3.5){{\footnotesize $\tt CIRS$}}
        \end{overpic}
    \end{center}
    \vspace{0.05in}
    \caption{Experimental results on monotone problems with 6-12 objects evaluating (1) success rate on finding a solution (left column) and (2) computation time (right column).}
    \label{fig:monotone_results}    
\end{figure}

{\bf Impact of Monotone Solver:} The efficiency of the monotone solver $\tt CIRS$ is evaluated on monotone problems with 6-12 objects given a time limitation of 3 minutes. Here "6 objects" corresponds to rearranging objects to be aligned in one row at the front of the workspace and 12 objects to occupy two rows. The metric for comparing monotone solutions involves success rate and computation time. 60 experiments are performed for each number of objects. Fig. \ref{fig:monotone_results} demonstrates that the proposed $\tt CIRS$ outperforms the complete alternatives $\tt mRS$ and $\tt \DFSDP$ with 57.3\% faster computation time (right column) on average. In harder cases (10 and 12 objects), the success rate (left column)  of the comparison points $\tt mRS$ and $\tt \DFSDP$ significantly drops while $\tt CIRS$ remains high (100\% for 10 objects and 88.7\% for 12 objects). It aligns with the observation that the proposed $\tt CIRS$ uses constraint reasoning to detect invalid actions ahead of time and performs online validity checking when deciding to move an object or not. Therefore, it saves significant time by not growing a tree, which will not lead to a solution. In contrast, $\tt mRS$ and $\tt \DFSDP$ do not perform any constraint reasoning and will explore many redundant branches of the search tree.

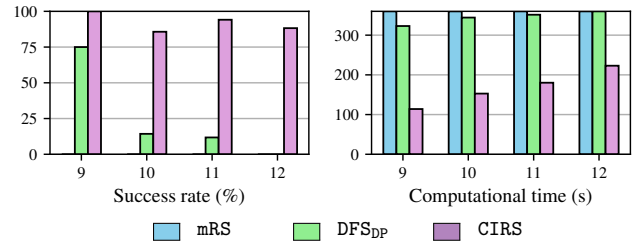
\begin{figure}[ht]
    \begin{center}
        \begin{overpic}[scale=0.68]{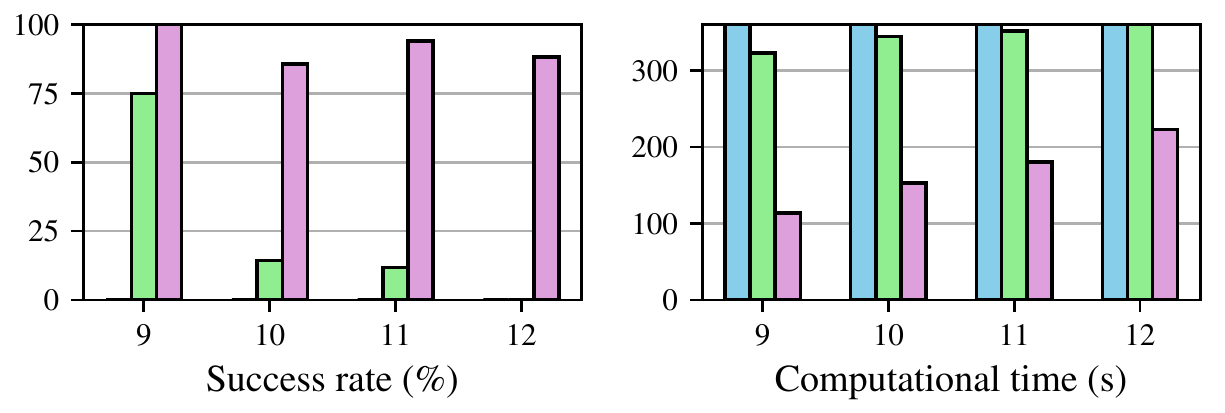}
        \definecolor{skyblue}{RGB}{135, 206, 235}
        \put(23,-3){\fcolorbox{black}{skyblue}{  }}
        \put(30, -3.5){{\footnotesize $\tt mRS$}}
        \definecolor{lightgreen}{RGB}{144, 238, 144}
        \put(45,-3){\fcolorbox{black}{lightgreen}{  }}
        \put(52, -3.5){{\footnotesize $\tt \DFSDP$}}
        \definecolor{plum}{RGB}{178, 132, 190}
        \put(67,-3){\fcolorbox{black}{plum}{  }}
        \put(74, -3.5){{\footnotesize $\tt CIRS$}}
        \end{overpic}
    \end{center}
    \vspace{0.05in}
    \caption{Experimental results on non-monotone problems with 9-12 objects evaluating (1) success rate on finding a solution (left column) and (2) computation time (right column).}
    \label{fig:non_monotone_results}    
\end{figure}

{\bf Performance in non-monotone problems:} The global task planner $\tt PERTS$ is evaluated on harder non-monotone instances with 9-12 objects given a time limitation of 6 minutes. $\tt CIRS$, $\tt \DFSDP$ and $\tt mRS$ are integrated as the local solvers in the $\tt PERTS$ structure for comparison. The metric for comparing non-monotone solutions involves success rate, computation time and the total number of actions to fulfill the tasks. 60 experiments are performed for each number of objects. Fig. \ref{fig:non_monotone_results} indicates how the efficiency of the local solver determines the success rate (left column) of solving non-monotone problems with the global task planner. $\tt PERTS(mRS)$ fails to solve non-monotone problems with at least 9 objects and the success rate of $\tt PERTS(\DFSDP)$ drops to 14\% in 10-object cases and 0\% in 12-object cases. Since the proposed $\tt CIRS$ is capable of growing trees much faster than comparison methods, $\tt PERTS(CIRS)$'s success rate remains high (91.7\%) and is on average 52.7\% faster than other methods in computational time (right column).


Table \ref{table:actions_non_monotone} provides the number of buffers needed to fulfill a non-monotone task together with the success rate of its variations of the $\tt PERTS$ planner. $\tt PERTS(CIRS)$ needs on average only 1.3 buffers to solve the harder non-monotone problems.  Given the $\tt PERTS$ global planner, even $\tt PERTS(\DFSDP)$ returns high-quality solutions with an average of 1.26 buffers needed when the approach can find a solution.

\section{Conclusion and Future Work}
\label{sec:conclusion}

This work improves the efficiency of monotone primitives for prehensile rearrangement in cluttered and confined spaces while maintaining properties. This is achieved by identifying the problem's combinatorial constraints to properly prune the search space. The primitive is integrated with a global planner to address non-monotone instances efficiently and with high-quality solutions. A useful pre-processing tool has been proposed for minimizing the cost of online collision checking in this domain via a labeled roadmap. The experiments demonstrate that the proposed integration achieves higher success rate, shorter computation time and uses fewer buffers to solve general prehensile rearrangement tasks.

An important extension involves considering the effects of perception. Occlusions may arise often in highly-constrained workspaces, which result in partial observability and uncertainty. These aspects can affect the decision-making process for rearranging objects, i.e., priority may be given to rearrange objects which can increase visibility of other objects, or for which the robot is more certain about their location. Non-prehensile actions can also be integrated with the proposed framework to provide even more effective algorithms that appropriately select between prehensile and non-prehensile actions. 



\newpage

\bibliographystyle{format/IEEEtran}
\bibliography{bib/c}

\end{document}